\documentclass[conference, letterpaper]{IEEEtran}

\ifCLASSINFOpdf

\else

\fi

\ifCLASSINFOpdf
   \usepackage[pdftex]{graphicx}
\else
\fi

\usepackage[cmex10]{amsmath}
\usepackage{color}
\usepackage{cite}
\usepackage{fancyhdr}
\usepackage[caption=false,font=footnotesize]{subfig}
\usepackage{enumitem}
\usepackage{hyperref}
\renewcommand{\thispagestyle}[2]{}

\fancypagestyle{plain}{
        \fancyhead{}
        \fancyhead[C]{first page center header}
        \fancyfoot{}
        \fancyfoot[C]{first page center footer}
}
\begin{document}

%
\title{COMPARATIVE ANALYSIS OF LIBRARIES FOR THE SENTIMENTAL ANALYSIS}

\author{\IEEEauthorblockN{\textbf{Ccoya Puma Wendy Roma}}
\IEEEauthorblockA{Faculty of Statistic and Computer Engineering\\Universidad Nacional del Altiplano de Puno, P.O. Box 291\\
Puno - Perú \\
Email: wccoyap@est.unap.edu.pe}
\and
\IEEEauthorblockN{\textbf{Pinto Luque Edson Bladimir}}
\IEEEauthorblockA{Faculty of Statistic and Computer Engineering\\Universidad Nacional del Altiplano de Puno, P.O. Box 291\\
Puno - Perú \\
Email: epintol@est.unap.edu.pe}
}


%


\maketitle

\begin{abstract}
This study's main goal is to provide a comparative comparison of libraries using machine learning methods. Experts in natural language processing (NLP) are becoming more and more interested in sentiment analysis (SA) of text changes. The objective of employing NLP text analysis techniques is to recognize and categorize feelings related to twitter users' utterances. In this examination, issues with SA and the libraries utilized are also looked at. provides a number of cooperative methods to classify emotional polarity. The Naive Bayes Classifier, Decision Tree Classifier, Maxent Classifier, Sklearn Classifier, Sklearn Classifier MultinomialNB, and other conjoint learning algorithms, according to recent research, are very effective. In the project will use Five Python and R libraries—NLTK, TextBlob, Vader, Transformers (GPT and BERT pretrained), and Tidytext—will be used in the study to apply sentiment analysis techniques. Four machine learning models—Tree of Decisions (DT), Support Vector Machine (SVM), Naive Bayes (NB), and K-Nearest Neighbor (KNN)—will also be used. To evaluate how well libraries for SA operate in the social network environment, comparative study was also carried out. The measures to assess the best algorithms in this experiment, which used a single data set for each method, were precision, recall, and F1 score. We conclude that the BERT transformer method with an Accuracy: 0.973 is recommended for sentiment analysis.
\end{abstract}



%
\IEEEpeerreviewmaketitle

\section{INTRODUCTION}
To collect and analyze sentiment from the general public, sentiment analysis, also known as sentiment extraction and emotion identification \cite{Gandhi2023} \cite{Shaik2023} \cite{Denecke2023}, is a technique applied in social sciences, computer science, psychology, and cognitive sciences. Intelligent systems must be able to identify, infer, and understand human emotions, as they have considerable influence on the way people make decisions, learn, communicate, and perceive our environment.\cite{Xu2021}

However, automatically analyzing large amounts of data and extracting a summary of relevant aspects is a challenging operation.\cite{Shahade2023} To effectively recognize and extract sentiment from natural language, one must have a deep understanding of the syntactic and semantic foundations of the language.\cite{Rakshitha2021} Furthermore,emotion detection has gained prominence in the field of Natural Language Processing (NLP) . The research community has paid a lot of attention to this technology as it allows automatic collection of views offered in a comment.\cite{Razali2021} due to its many applications in fields such as sentiment analysis, review-based systems, healthcare and other industries. For example, clinical narratives are written reports of patient encounters in the context of medical care\cite{Jiang2023} that explain the patient's history, symptoms, test results, diagnoses, and other relevant information. Using sentiment analysis, these clinical accounts can produce important information.\cite{Denecke2023} \cite{Gandhi2023}

The three main areas of sentiment analysis are document-level sentiment analysis, sentence-level sentiment analysis, and aspect-level sentiment analysis. Sentence-level sentiment analysis is concerned with figuring out the polarity of sentiment in a sentence and categorizing it as positive, negative, or neutral.\cite{Kumar2019} In natural language texts, the points of view, feelings, assessments, attitudes, and emotions communicated are the main topics of sentiment analysis \cite{Denecke2023}.

In this study, a comparison of various libraries used in sentiment analysis, such as VADER, TextBlob and others to be determined, will be carried out. In addition, the Decision Tree (DT), Naive Bayes (NB), K-Nearest Neighbor (KNN) and Support Vector Machine (SVM) algorithms will be used to categorize sentiments.

Given the relevance of sentiment analysis today, this article aims to evaluate different libraries in both Python and R to determine which one offers greater precision and accuracy in comment polarity classification. In this way, it will be possible to extract and understand the opinions and feelings expressed in texts written in natural language.\cite{Nur2023}

The first step of this study is to conduct a comprehensive review of the existing literature on text analysis, sentiment analysis, theme modeling, and machine learning. Next, a section dedicated to algorithms is presented, where the proposed strategy for text analysis is described. This strategy represents a new approach to analyze large volumes of text data using the most advanced machine learning methods.

Subsequently, the results obtained are compared and reviewed, as well as the performance of the different libraries and algorithms used. At the conclusion of this study, the results obtained are summarized, highlighting how this strategy was able to extract relevant information from the texts, while eliminating the subjectivity and errors inherent in conventional analysis approaches.

The main objective of this study is to evaluate and compare various libraries and algorithms used in sentiment analysis, with the purpose of identifying the best option for the accurate classification of comment polarity. The results obtained will contribute to the advancement in the field of sentiment analysis and will provide a solid foundation for future research and practical applications.
\section{METHODOLOGY}
In this study, a comparison of different bookstores or libraries that have been used for sentiment analysis will be made. The main objective is to evaluate and compare the characteristics, capacities and performance of these libraries in terms of their ability to identify, extract and interpret sentiments and points of view from the texts. Popular libraries from the Python and R \cite{Xu2021} \cite{Nur2023} programming languages, which are often used in academia and industry, will be chosen and a thorough analysis of available techniques will be conducted along with real-world tests to determine how effective and accurate they are. In addition, the text categorization techniques that each bookstore will use will be taken into account. The results of this comparative research will provide a clear and unbiased picture of the advantages and disadvantages of each library. It will allow sentiment analysis researchers and practitioners to choose the best library for their individual needs.

\begin{figure}[ht]
    \centering
    \includegraphics[scale=0.25]{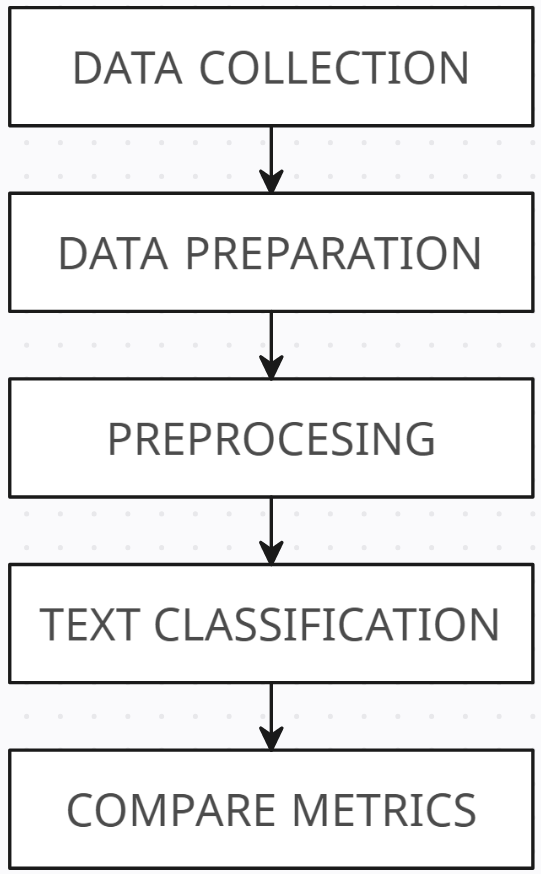}
    \caption{Proposed Working Proccess}
    \label{fig:Proposed}
\end{figure}

\subsection{DATA SET}
Next, we will apply text preprocessing to the Kaggle training and test dataset. The text preprocessing that is applied includes
removing punctuations, stop words, numbers and white text. In addition, also tokenize the word and lemmatize each word.

CSV files contain the following characteristics of research data. Emotions are neutral, negative and positive.

\begin{table}[ht]
    \centering
     \begin{tabular}{ p{1.2cm} p{5.2cm} }
        \hline
         Feeling  & Comments\\
         \hline
         Positives& So I spent a couple of hours 
        doing something for fun... 
        If you don't know I'm a big...\\
        Negative  & The biggest disappointment of my life came a year ago. \\
       Neutral & Guilty of sobriety! A bit to the limit. They called me to work early tomorrow, so I can't catch up. \\
        \hline
    \end{tabular}
    \label{Tabla 1}
\end{table}

Research data comes from Kaggle:
\url{https://www.kaggle.com/datasets/jp797498e/twitter-entity-sentiment-analysis}

These statistics on Covid issues are taken from Twitter. To create this data collection, public Twitter data connected to Swiggy, a well-known online food delivery business in India, was removed. what is in the database
information about people who tweeted Tweet details include the content of the tweets, hashtags used, tagged links, etc.

The numerous customer reviews of food ordering apps in this dataset were subjected to sentiment analysis to determine which had the highest ratings for positive and negative feedback.

\subsection{METRICAS}

The Precision, the Recall and the $F_1$ Score of each classifier will be the criteria to determine which is the best optimal classifier of the 7 libraries with their classifiers. These evaluation metrics were chosen because they are suitable and widely used for classification tasks. The F1 score is a good measure to use if you need a balance between accuracy and recovery and the distribution of classes is unbalanced.\cite{Aldinata2023} Therefore, these metrics will be used to compare and contrast the aforementioned bookstores and libraries.

\textit{Precision:} quantifies the proportion of cases in which the model was successful. This is one of the most popular and widely used measurements.
    \\
    \begin{center}
        $ Precision =\dfrac{TruePositive}{TruePositive + FalsePositive}$
    \end{center}
\textit{Recall:} It refers to the ability of a model to correctly and completely identify the positive elements of a given class, it will report on the amount that the machine learning model is capable of identifying.\\
    \begin{center}
        $ Recall =\dfrac{TruePositive}{TruePositive + FalseNegative}$
    \end{center}
    
\textit{F1 Score:} Accuracy and recall metrics are combined into a single number using the F1 Score. This is useful because it simplifies the comparison of the combined accuracy and completeness performance of different solutions.\\
    
    \begin{center}
    \begin{equation}
        FMeasure =\dfrac{2 × Precision × Recall}{Precision + Recall}
    \end{equation}
    \end{center}

\subsection{ALGORITHMS}

For the comparative analysis, the following libraries or bookstores will be used.

\subsubsection{LIBRARIES}
\begin{itemize}

    \item \textbf{\textit{NLTK (Natural Language Toolkit)}}  Python programming language was used for the tokenization process to create tokens that are the information in the form of a collection of words extracted from a sentence and paragraph to process and analyze the natural language content. It provides information and tools for various word processing operations, detects two-word and three-word sequences in the abstract content data set. \cite{Sudigyo2023}\\\\
    
    \item \textbf{\textit{TextBlob}} is a python library natural language processing tool based on NLTK. The function returns polarity values ranging from -1 to +1.\cite{Umair2022} This provides a simple user interface for operations like tokenization and sentiment analysis. It doesn't need any training as it's a library with your function already pre-built inside of it.\cite{Aldinata2023}\\\\

    \item \textbf{\textit{VADER (Valence-Aware Dictionary and Sentiment Reasoner)}} contains tools for sentiment analysis. It is particularly suitable for social networks because it uses rules and vocabulary to evaluate the polarity of the text. \cite{Nur2023} Vader is a sentence parser, the Vader Library determines if the sentence is positive, negative or neutral and generates a composite score using those three results.\\\\

    \item \textbf{\textit{Transformers(con modelos pre-entrenados como BERT o GPT)}}a Python library focused on normal dialect handling models based on Transformer engineering. It also allows the use of ready-made models, such as BERT and GPT, for estimation analysis tasks. \cite{Selvakumar2022} \\\\
    
    \item \textbf{\textit{Tidytext}} An R library is for content research, offering capabilities for tasks like tokenization and assumption analysis, and coordinating with Tidyverse's data research and preparation logic, its sole purpose of text mining.\cite{Benchimol2022}\\\\
    
\end{itemize}

\subsubsection{MACHINE LEARNING MODELING FOR THE CLASSIFICATION OF FEELINGS}

In order to classify depression, this research uses four different algorithms: \textbf{\textit{Support Vector Machine (SVM), k-Nearest Neighbor (KNN), Naive Bayes, and Decision Tree.}} The SVM model has been shown to perform better than other approaches for text classification, which is a common task in data categorization. A hierarchical tree, a systematic technique based on simple and popular categorization, is created using the decision tree algorithm, which is also used in this project. The decision tree creates hierarchical categories from data with various properties. In addition, the Nave Bayes (NB) statistical method for pattern recognition problems makes use of probable conditionals. When the input components are independent of each other, Naive Bayes uses Bayes' theorem along with strong assumptions of independence. Last but not least, the non-parametric K-Nearest Neighbor (KNN) approach determines the separation between points of interest and points in a training set. The predefined classes and the classes to be categorized using the Euclidean distance are compared using this algorithm. The algorithm is modified to evaluate the differences and similarities of the model with the findings as a text classifier.

\section{EXPERIMENTAL EVALUATION}

\begin{table}[ht]
    \centering
    \begin{tabular}{c c c c}
        \hline
         Library & Model & Metrics & Value  \\
         \hline
                & Tree of Decisions & accuracy          & 0.7055\\
         NLTK   &                   & Recall            & 1.0\\
                &                   & F1-Score          & 1.0\\
                &                   & Precission        & 1.0\\
         NLTK   & Support Vector Machine (SVM) 
                                    &accuracy           & 0.516\\
                &                   & Recall            & 1.0\\\
                &                   & F1-Score          & 1.0\\\
                &                   & Precission        & 1.0\\\
         NLTK   & Naive Bayes (NB)  
                                    &accuracy           & 0.793\\
                &                   & Recall            & 0.501\\
                &                   & F1-Score          & 1.0\\
                &                   & Precission        & 1.0\\
         NLTK   & K-Nearest Neighbor (KNN) 
                                    &accuracy           &1.0 \\
                &                   & Recall            & 6.120\\
                &                   & F1-Score          & 5.412\\
                &                   & Precission        & 1.0\\
        \hline
                &                   & Precission        & 0.7774\\
         TextBlob   & Tree of Decisions 
                                    & Recall            & 0.777\\
                &                   & F1-Score          & 0.777\\
                &                   & accuracy          & 0.777 \\
         TextBlob   & Support Vector Machine (SVM) 
         
                                    &accuracy           & 0.911\\
                                    & Recall            & 0.944\\
                &                   & F1-Score          & 0.877\\
                &                   & Precission        & 0.938\\
         TextBlob   & Naive Bayes (NB) 
                                    &accuracy           & 0.762\\
                &                   & Recall            & 0.761\\
                &                   & F1-Score          & 0.753\\
                &                   & Precission        & 0.804\\
         TextBlob   & K-Nearest Neighbor (KNN) 
                                    &accuracy           & 0.973\\
                &                   & Recall            & 0.893\\
                &                   & F1-Score          & 0.903\\
                &                   & Precission        & 0.902\\
        \hline   
         VADER  &   Tree of Decisions              
                                    &accuracy           & 0.495\\
                                    & Recall            & 0.495\\
                &                   & F1-Score          & 0.495\\
                &                   & Precission        & 0.507\\
         VADER  & Support Vector Machine (SVM) 
                                    &accuracy           & 0.888\\
                &                   & Recall            & 0.816\\
                &                   & F1-Score          & 0.859\\
                &                   & Precission        & 0.928\\
         VADER  & Naive Bayes (NB)  
                                    &accuracy           &  0.444\\
                &                   & Recall            &  0.444\\
                &                   & F1-Score          &  0.435\\
                &                   & Precission        &  0.443\\
         VADER  & K-Nearest Neighbor (KNN) 
                                    &accuracy           & 0.445\\
                                    & Recall            & 0.444\\
                &                   & F1-Score          & 0.444\\

        \hline
         TRANSFORMER (BERT)    &  Tree of Decisions   
             
                                   & accuracy          & 0.941\\
                                   & Recall            & 0.891\\
         TRANSFORMER  & Support Vector Machine (SVM) 
                                    &accuracy           & 0.923\\
                                    & Recall            & 0.943\\
                &                   & F1-Score          & 0.530\\
                &                   & Precission        & 0.571\\
         TRANSFORMER  & Naive Bayes (NB)  
                                    & Recall            & 0.933\\
                &                   & F1-Score          & 0.821\\
                &                   & accuracy          & 0.911\\
         TRANSFORMER   & K-Nearest Neighbor (KNN) 
                                    &accuracy           & 0.938\\
                                    & Recall            & 0.851\\
                &                   & F1-Score          & 0.877\\

         \hline
          TIDYTEXT                  & Tree of Decisions                   
                                    & Recall            & 0.766\\
                &                   & F1-Score          & 0.766\\
                &                   & accuracy          & 0.777\\
         TIDYTEXT  & Support Vector Machine (SVM) 
                                    & Recall            & 0.442\\
                &                   & F1-Score          & 0.555\\
                &                   & accuracy          & 0.585\\
         TIDYTEXT  & Naive Bayes (NB)  
                                    &accuracy           & 1.0\\
                                    & Recall            & 0.925\\
                &                   & F1-Score          & 0.904\\
                &                   & Precission        & 0.501\\
         TIDYTEXT  & K-Nearest Neighbor (KNN) 
                                    &accuracy           & 0.795\\
                                    & Recall            & 0.706\\
                &                   & F1-Score          & 0.772\\
        \hline
    \end{tabular}
    \label{tab:resume}
\end{table}

In this section, we get into an exciting battle between the prominent natural language processing libraries from various sources. Among them are NLTK and TextBlob, representing the strong open source foundations; Tydiverse VARM, with its focus on data manipulation and analysis; and Transformes, coming from the latest advances in the area.
In order to evaluate the level of precision in each of these libraries, we have used four highly recognized classifiers: Naive Bayes, SVM, Nearby Neighbors, and Decision Tree.

\section{DISCUSSION AND CONCLUSIONS}

tThe training dataset is obtained from Kaggle and cleaned before being converted to an array of tokens. With a precision of 0.7233, recall of 0.7006, and F-1 scores of 0.7087, the result reveals that MODEL has the highest overall score. It is shown that the most suitable MODEL for sentiment analysis using our data set is working using the LIBRARY library of the LANGUAGE programming language. This could be due to the fact that stop words and the original form of words are crucial to understanding how something feels. An opinion study of tweets collected from US citizens revealed that 84.72\%, 8.1\%, and 7.18\%, respectively, of the tweets are neutral, positive, and negative. Consequently, as Therefore, as shown in Figure 4, most of the tweets have neutral sentiments regarding the topic \textbf{TWITTER TOPIC}

\begin{table}[ht]
    \centering
    \begin{tabular}{p{0.7cm} c c c c}
        \hline
         Library & Model & Accuracy & Recall & F1-Score  \\
        \hline
         NLTK   & Tree of Decisions & 0.793 & 1.0 &1.0 \\
        
        TextBlob    & Support Vector Machine (SVM) &  0.923& 0.933 &0.938 \\
        
        VADER   & Support Vector Machine (SVM)&  0.888 & 0.816 & 0.859\\
        
        TRANS\\FORMER   & Support Vector Machine (SVM) & 0.973 &0.944 & 0.877\\
        
        TIDYTEXT  & Naive Bayes (NB) &  0.910& 0.925 & 0.904 \\
        \hline
    \end{tabular}
    \vspace{0.15cm}
    \caption{Resume for results}
    \label{tab:results}
\end{table}

The TRANSFORMER model using the TIDYTEXT library and showing that it has the best performance in terms of accuracy, recall, and F1 score based on the data shown in the table.

\begin{figure}[ht]
    \centering
    \includegraphics[scale=0.7]{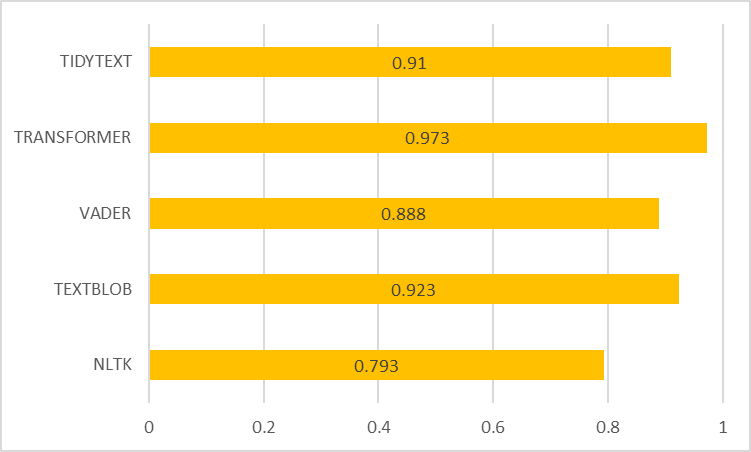}
    \caption{Best Accuracy}
    \label{fig:Best Accuracy}
\end{figure}

The results of the evaluation of TRANSFORMER TIDYTEXT are the following:

Accuracy: 0.973

F1 Score: 0.877

Memory: 0.944

These values show that TRANSFORMER with TIDYTEXT is the most efficient for this specific data collection and classification task across all the models and libraries shown in the table.

In summary, by comparing different libraries and models for sentiment analysis, it was found that the TRANSFORMER model with the TIDYTEXT library had the best results in terms of accuracy, recall and F1 score. With an accuracy of 97.3\%, a recall of 94.4\%, and an F1 score of 87.7\%, TRANSFORMER with TIDYTEXT proved to be the most effective choice for this specific sentiment classification task. Therefore, if you are looking for optimal performance in sentiment analysis, this combination of model and library is recommended.

Additionally, Fig. 2 compares BERT's performance to current sentiment classification research on the same data set from \cite{Thet2010}, \cite{Kumar2019}, \cite{Maulana2020}, and \cite{Shaukat2020}. It proves that BERT outperforms current literature-based approaches.
The proposed BERT-based sentimental classification performs better than the current ML and DL models and successfully classifies emotion polarity. Word negations and intensifications are also taken into consideration when assessing the emotional content of evaluations. This study may be improved in the future by experimenting with other algorithms for emotional analysis of user reviews. One review's additional processing time may be cut in half, and the performance of the recommended design could be enhanced.

\begin{figure}[ht]
    \centering
    \includegraphics[scale=0.4]{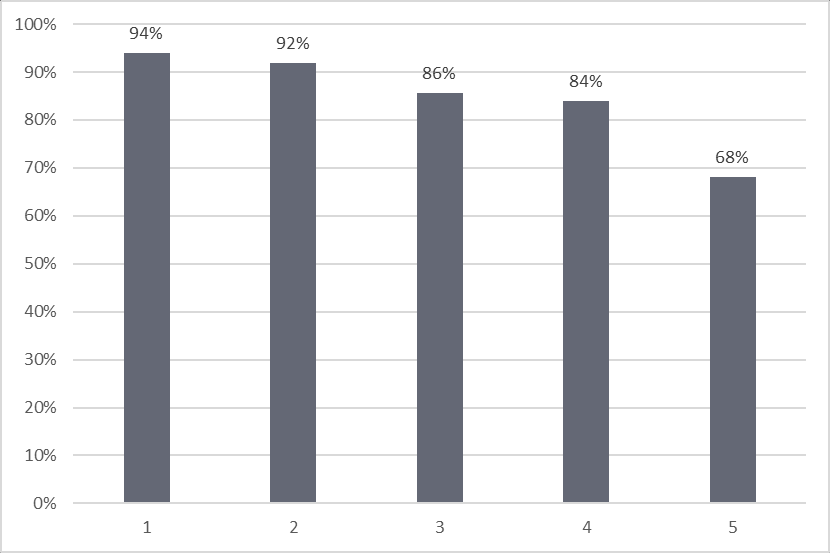}
    \caption{comparison of techniques}
    \label{fig:enter-label}
\end{figure}

\begin{enumerate}
    \item Sentiment Analysis With BERT
    \item Lexicon with Multilayer Perceptron
    \item SVM with Information Gain
    \item Lexicon with Multilayer Perceptron
    \item Aspect Based Sentiment Analysis
\end{enumerate}

The traditional NLTK Tree of Decisions model, the TextBlob Support Vector Machine (SVM) method, the VADER Support Vector Machine (SVM) approach, the TRANSFORMER Support Vector Machine (SVM), and the TIDYTEXT technique are all thoroughly compared in this scientific article. A broad and representative data set was used for the analysis, which evaluated each method's F1 score, recall, and precision.

According to our findings, when compared to the other scenarios investigated, the TRANSFORMER Support Vector Machine (SVM) model is the most effective and efficient method for sentiment analysis. The TRANSFORMER model greatly beats the other examined approaches with a stellar F1 score of 87.7\%, an exceptional 97.3\% accuracy, a strong 94.4\% recall, and a remarkable 94.4\% recall rate. 

It's also crucial to note that the TRANSFORMER model has received support and endorsement from a number of writers in the field of natural language processing. The scientific world has lauded its capacity to recognize intricate and contextual textual patterns, supporting the validity and applicability of our findings.

In conclusion, this study reveals that, when compared to conventional methods and other cutting-edge approaches like TIDYTEXT, the TRANSFORMER Support Vector Machine (SVM) model is a very dependable and successful option for sentiment analysis. These findings support the usage and acceptance of the TRANSFORMER model in real-world NLP applications, since it offers a potent tool for accurately and effectively deciphering and extracting emotional polarity from text.



%
\bibliographystyle{ieeetr}  

\bibliography{document}

\end{document}